\documentclass[sigconf]{acmart}
\usepackage{booktabs} % For formal tables

\copyrightyear{2018}
\acmYear{2018}
\setcopyright{acmcopyright}
\acmConference[WSDM 2018]{WSDM 2018: The Eleventh ACM International
Conference on Web Search and Data Mining }{February 5--9, 2018}{Marina
Del Rey, CA, USA}
\acmPrice{15.00}
\acmDOI{10.1145/3159652.3159705}
\acmISBN{978-1-4503-5581-0/18/02}

\begin{document}
\title{Who Will Share My Image? Predicting the Content Diffusion Path in Online Social Networks}

\author{Wenjian Hu}
\affiliation{%
  \institution{University of California, Davis}
}
\email{wjhu@ucdavis.edu}

\author{Krishna Kumar Singh}
\authornote{These two authors contributed equally.}
\affiliation{%
  \institution{University of California, Davis}
}
\email{krsingh@ucdavis.edu}

\author{Fanyi Xiao\footnotemark[1]}
%\authornote{These two authors contributed equally.}
\affiliation{%
  \institution{University of California, Davis}
}
\email{fanyix@cs.ucdavis.edu}

\author{Jinyoung Han}
\affiliation{%
  \institution{Hanyang University}
}
\email{jinyounghan@hanyang.ac.kr}

\author{Chen-Nee Chuah}
\affiliation{%
  \institution{University of California, Davis}
}
\email{chuah@ucdavis.edu}

\author{Yong Jae Lee}
\affiliation{%
  \institution{University of California, Davis}
}
\email{yongjaelee@ucdavis.edu}

\renewcommand{\shortauthors}{W. Hu et al.}

\begin{abstract}
Content popularity prediction has been extensively studied due to its importance and interest for both users and hosts of social media sites like Facebook, Instagram, Twitter, and Pinterest. However, existing work mainly focuses on modeling popularity using a single metric such as the total number of likes or shares. In this work, we propose Diffusion-LSTM, a memory-based deep recurrent network that learns to recursively predict the entire diffusion path of an image through a social network. By combining user social features and image features, and encoding the diffusion path taken thus far with an explicit memory cell, our model predicts the diffusion path of an image more accurately compared to alternate baselines that either encode only image or social features, or lack memory. By mapping individual users to user prototypes, our model can generalize to new users not seen during training. Finally, we demonstrate our model's capability of generating diffusion trees, and show that the generated trees closely resemble ground-truth trees.
%Predicting the popularity of content is important and intriguing for both users and hosts of social media sites, such as Facebook, Google+, Instagram, Twitter, and Pinterest. Existing approaches for popularity prediction have largely focused on predicting a single metric, like the total number of comments, likes or shares of posts. We instead propose to learn and predict the \emph{entire diffusion path} of an image in a social network. To this end, we design a tree-structured long short-term memory (LSTM) network, dubbed as Diffusion-LSTM. By combining user social features and image features together with the encoded diffusion path history stored in an explicit memory cell, our Diffusion-LSTM is able to keep track of the posting history of an image and predicts its diffusion path better than alternate baselines that rely only on either image or social features, or do not encode the posting history. Our model generalizes to new users who are not included in the training set, through a mapping between individual users and user prototypes. Finally, we also demonstrate that our Diffusion-LSTM can \emph{generate} meaningful diffusion trees that closely resemble ground-truth trees.
\end{abstract}
%{\color{red} and have similar characteristics.}

\begin{CCSXML}
<ccs2012>
<concept>
<concept_id>10003033.10003106.10003114.10011730</concept_id>
<concept_desc>Networks~Online social networks</concept_desc>
<concept_significance>500</concept_significance>
</concept>
<concept>
<concept_id>10003120.10003130.10003134.10003293</concept_id>
<concept_desc>Human-centered computing~Social network analysis</concept_desc>
<concept_significance>500</concept_significance>
</concept>
<concept>
<concept_id>10010147.10010257.10010258.10010259.10010263</concept_id>
<concept_desc>Computing methodologies~Supervised learning by classification</concept_desc>
<concept_significance>500</concept_significance>
</concept>
<concept>
<concept_id>10002951.10003227.10003351</concept_id>
<concept_desc>Information systems~Data mining</concept_desc>
<concept_significance>300</concept_significance>
</concept>
<concept>
<concept_id>10010147.10010257.10010293.10010294</concept_id>
<concept_desc>Computing methodologies~Neural networks</concept_desc>
<concept_significance>300</concept_significance>
</concept>
</ccs2012>
\end{CCSXML}

\ccsdesc[500]{Networks~Online social networks}
\ccsdesc[500]{Human-centered computing~Social network analysis}
\ccsdesc[500]{Computing methodologies~Supervised learning by classification}
\ccsdesc[300]{Information systems~Data mining}
\ccsdesc[300]{Computing methodologies~Neural networks}

\keywords{Online Social Networks; Recurrent Neural Networks; Diffusion Path Prediction}

\maketitle
\section{Introduction}

\begin{figure*}
\includegraphics[width=0.9\linewidth]{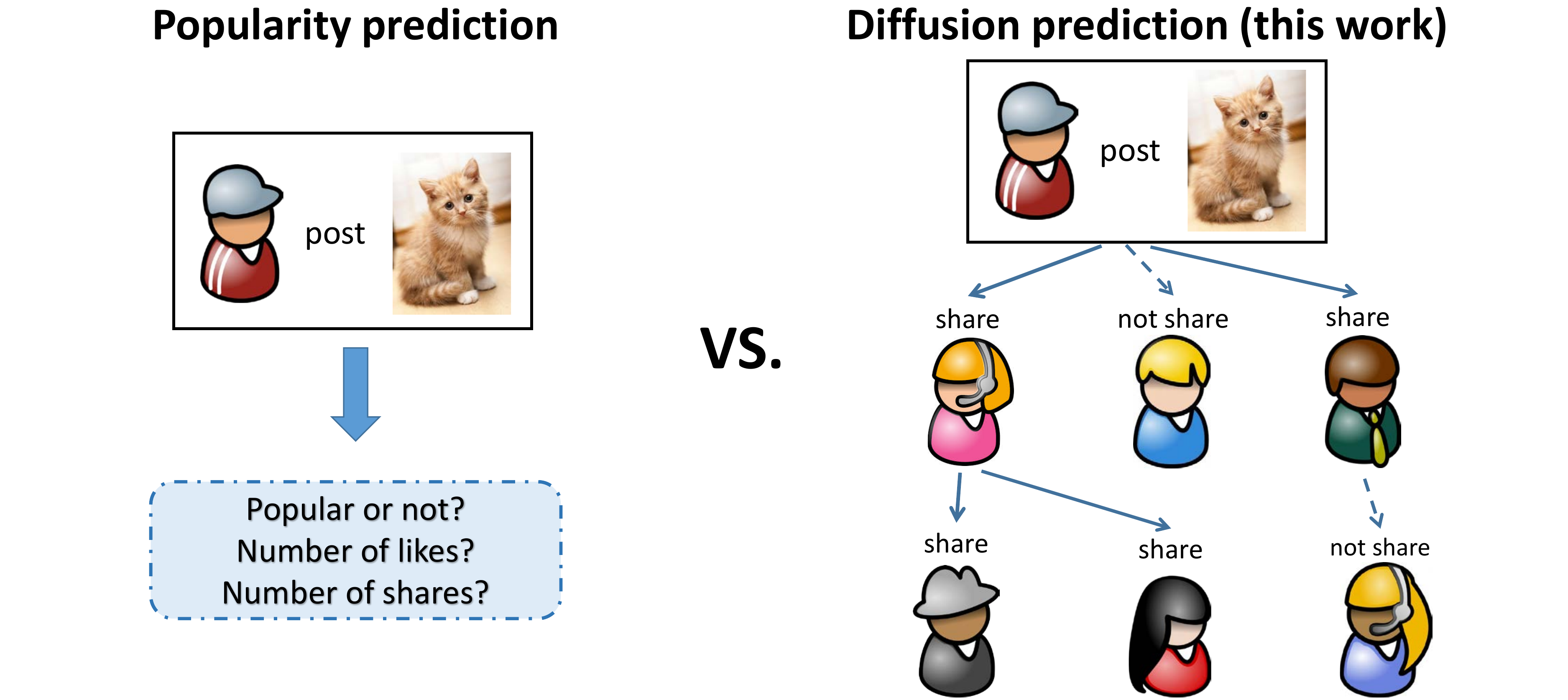}
  \centering
  \caption{\textbf{(Left)}: Prior work on popularity prediction predicts whether a user's post will become popular or not as measured by a single metric such as the total number of users who liked or shared the post. \textbf{(Right)}: Our work on diffusion prediction predicts the entire propagation path of the user's post.  Specifically, our approach recursively predicts which of the user's friends will share/re-share the post using the image features, user's social features, and the previous sharing history.}
  \label{fig:LSTM_overview}
\end{figure*}

Online social networks (OSNs) have seen phenomenal growth: over a billion users on Facebook, and hundreds of millions on Google+, Twitter, and Pinterest. This new ecosystem of content generation and sharing enables rich data to be collected and analyzed to study individual and group behavior, as well as to study how information is diffused over social graphs. In particular, viral diffusion (also known as ``word-of-mouth'' diffusion)~\cite{yang-icml2013, kimura-ecpdmkd2006, bakshy-www2012} has been shown to be an important mechanism for advertising a new idea, technology, content, or product. Unlike the mass broadcast counterpart in which a single user directly spreads information to most recipients, in viral diffusion many
individuals participate in spreading information in a chain-like structure. %{\color{red}Despite the large theoretical and empirical literature on information diffusion~\cite{kimura-ecpdmkd2006,bakshy-www2012} in online social networks, there has been relatively little work about the viral diffusion.}

In this paper, we are interested in modeling how image content gets propagated through a social network. We focus on images because it has become one of the most important containers for sharing thoughts, interests, and feelings in OSNs (e.g., Pinterest, Instagram, Tumblr, Imgur).  Unlike existing work~\cite{huberman-arxiv2008,agarwal-is2010,li-icikm2013,khosla-www2014,gelli-mm2015,deza-cvpr2015,borges-icpmdase2016}, which largely focus on a single absolute metric such as the total number of users who liked/shared a content (see Figure~\ref{fig:LSTM_overview} left), we are instead interested in modeling the \emph{entire structure of the content propagation path} (see Figure~\ref{fig:LSTM_overview} right). To this end, we propose Diffusion-LSTM, a novel memory-based deep network that recursively predicts each step of the image content diffusion path, using both the
history of social context features of all previous users in the diffusion path as well as the image content features.  
%%such as predicting which of his/her friends will share the post based on the current social and image features,
%or just predicting immediate friends of the content generator who will share the content further
%\yj{link prediction works [20-22], which infer new interactions among its members given a snapshot of a social network}
%(see Fig.~\ref{fig:LSTM_overview} center)

%\yj{We may need to highlight the technical contribution a bit more.  Also, maybe explain the challenges and why we need prototypes.}

Apart from an image's content, the history of users in the diffusion path plays a key role in deciding whether the image will be further propagated, and if so, to whom next. For example, if we only know that the image was shared by a user who likes soccer, then the likelihood that the image will be shared by a soccer fan will be high.  However, if we also know that the image was initially shared by a tennis fan before being shared by the soccer fan, then there is a high-likelihood that the next user will like either soccer, tennis, or sports in general.  Similarly, if an image is initially shared by an influential user with lots of followers, then its likelihood to be propagated will remain high even after being shared by a user with only a few followers. Hence, we design our Diffusion-LSTM to capture the history of the social characteristics of all of the previous users who have shared the image in order to predict its future propagation path.  Furthermore, in order to generalize our prediction model to new unseen users, we create prototype users by grouping individual users according to their social features. %{\color{red} The main challenges of this work include: (1) how to correctly model the propagation history of an image through a social network; (2) how to deal with unseen users; and (3) how to properly combine a user's social information with the image content features.}
%the unbalanced user dataset and 
%So, knowing the history of the users in the propagation path is critical for correct prediction of the image propagation.

%[CHOOSE ONE EXAMPLE]

To the best of our knowledge, our work is the first attempt to generate the complete diffusion path of an image through an online social network.  Our results show superior diffusion path prediction performance compared to alternative baselines that lack history information or rely only on either image or social information.  Finally, while we use Pinterest data to evaluate our model, we design it to be general and applicable to other image-driven social media as well.

%Our Diffusion-LSTM network accounts for the history of users who have shared the image in order to predict the image's future propagation path.

%We demonstrate our approach on \emph{interest-driven Pinterest data}, but we design it to be applicable to other image-driven social media. %We believe this work can provide an empirical ground for developing efficient advertisement strategies for business products and for social good, and resource management approaches in sharing content on social media as well as mobile social applications (via cell phones or tablets).

%It is worth noting that patterns of image propagation would be different across different types of social media (friendship-based e.g., Facebook vs. interest-driven e.g., Pinterest), networks (static vs. mobile), or use cases (e.g., malicious vs. benign).

%\textbf{(Center)}: %Prior work on link prediction
  %\kr{Prior work on link prediction predicts which of the poster's friends will share the post based on the current social and image features.} 
\section{Related Work}

A large number of recent efforts have explored ways to predict content popularity, including for images~\cite{khosla-www2014,totti-cws2014,cheng-www2014,gelli-mm2015,deza-cvpr2015,guerini-icsc2015}, videos~\cite{li-icikm2013}, GitHub repositories~\cite{borges-icpmdase2016}, blogs~\cite{agarwal-is2010}, memes~\cite{weng-sr2013}, and tweets~\cite{huberman-arxiv2008,hong-www2011,kupavskii-aaai2013,ma-jasist2013}, by combing content features with user social features. In contrast to these prior work, which mainly focus on predicting a popularity score (e.g., number of shares) of the content, we aim to predict the entire content diffusion path through the social network, which is a much more challenging task.  Predicting not only how many users will share a content but also the characteristics of each sharer in the diffusion sequence would enable deeper understanding of the diffusion process; e.g., one could identify the critical users in the diffusion path~\cite{kimura-ecpdmkd2006,bakshy-www2012} which would be useful e.g., for targeted advertisements.

Link prediction algorithms, which predict future edges between the nodes by using the current state of the graph~\cite{nowell-jaist2007,wang-kdd2011,scellato-kdd2011,backstorm-wsdm2011,grover-kdd2016} are also related. Our approach is different in that it \emph{recursively} predicts the entire diffusion path of a content (an image) using both the content and users' social characteristics rather than predicting future edges for an existing graph.  %Also, the diffusion path is dependent on the content which can be new/unseen and none of the exiting diffusion paths in training will correspond to it directly to perform link prediction.
%\kr{There are also some works on link prediction~\cite{nowell-jaist2007,wang-kdd2011,scellato-kdd2011} in which they predict future edges between the nodes by using the current state of the graph. Where as in our approach for a given content (like image), we predict its whole diffusion path using its content and history of previous users' social context feature rather using the entire tree/graph. Also, the edges between the users are dependent on the propagated content and can vary for different contents. [NOT SURE, IT CAN BE TRUE FOR SOME SPECIAL CASE OF LINK PREDICTION]}

Our Diffusion-LSTM architecture for social network tree prediction is related to Tree-LSTMs~\cite{tai-acl2015,zhu-icml2015,zhang-naacl2016} in natural language processing (NLP).  Unlike standard linear chain LSTMs, a Tree-LSTM takes the hidden vectors from multiple children to predict the next hidden state, and has shown superior performance for various NLP tasks like sentiment prediction and sentence semantic relatedness. However, due to their bottom-up structure~\cite{tai-acl2015,zhu-icml2015}, they are not well suited for predicting/generating the content diffusion path in a social network, since the content diffusion path prediction has to start from the root user. Recent work by~\cite{zhang-naacl2016} predicts the tree structure in a top-down manner, but it uses techniques that are specific for language dependency tree prediction, and thus cannot be directly used for social network tree prediction. In contrast to these works, our Diffusion-LSTM generates the content diffusion path in a top-down manner for social network tree prediction. To the best of our knowledge, our work is the first attempt to generate the complete diffusion path of an image through a social network. % it can be easily generalized to predict other social network structures.

Finally, our work is also related to image captioning in computer vision.  In image captioning (e.g.,~\cite{kapathy-cvpr2015,chen-cvpr2015,donahue-cvpr2015,vinyals-cvpr2015,johnson-cvpr2016}), an LSTM takes the image semantic feature~\cite{simonyan-iclr2015,krizhevsky-nips2012} along with the history of previous words to recursively predict the next word. In contrast, in our method (Diffusion-LSTM), an LSTM takes the image semantic feature along with the history of previous users' social context features to recursively predict the image propagation path taken through the social network.

%\kr{LSTM has been previousily used on image for image captioning~\cite{kapathy-cvpr2015,chen-cvpr2015,donahue-cvpr2015,vinyals-cvpr2015} and visual question answering (VQA)~\cite{antol-iccv2015,lu-nips2015}. For image captioning, LSTM takes image semantic feature~\cite{simonyan-iclr2015,krizhevsky-nips2012} along with the history of previous words to recursively predict the next word. Whereas in our work LSTM takes the history of previous users' social context feature along with image semantic feature to predict the image propagation through the social network tree.  }
%In this work, we  will use the VGG 16-layer network [17], pretrained on the ImageNet [18] dataset, to extract $4096$ dimensions image features for images as a part of data preprocessing. Since AlexNet [19] won the ImageNet $2012$ competition, there has lots of interest and work toward deep convolutional models. The VGG 16-layer network is one of the most popular model used to extract representative image features. 
\begin{figure*}
\includegraphics[width=1\linewidth]{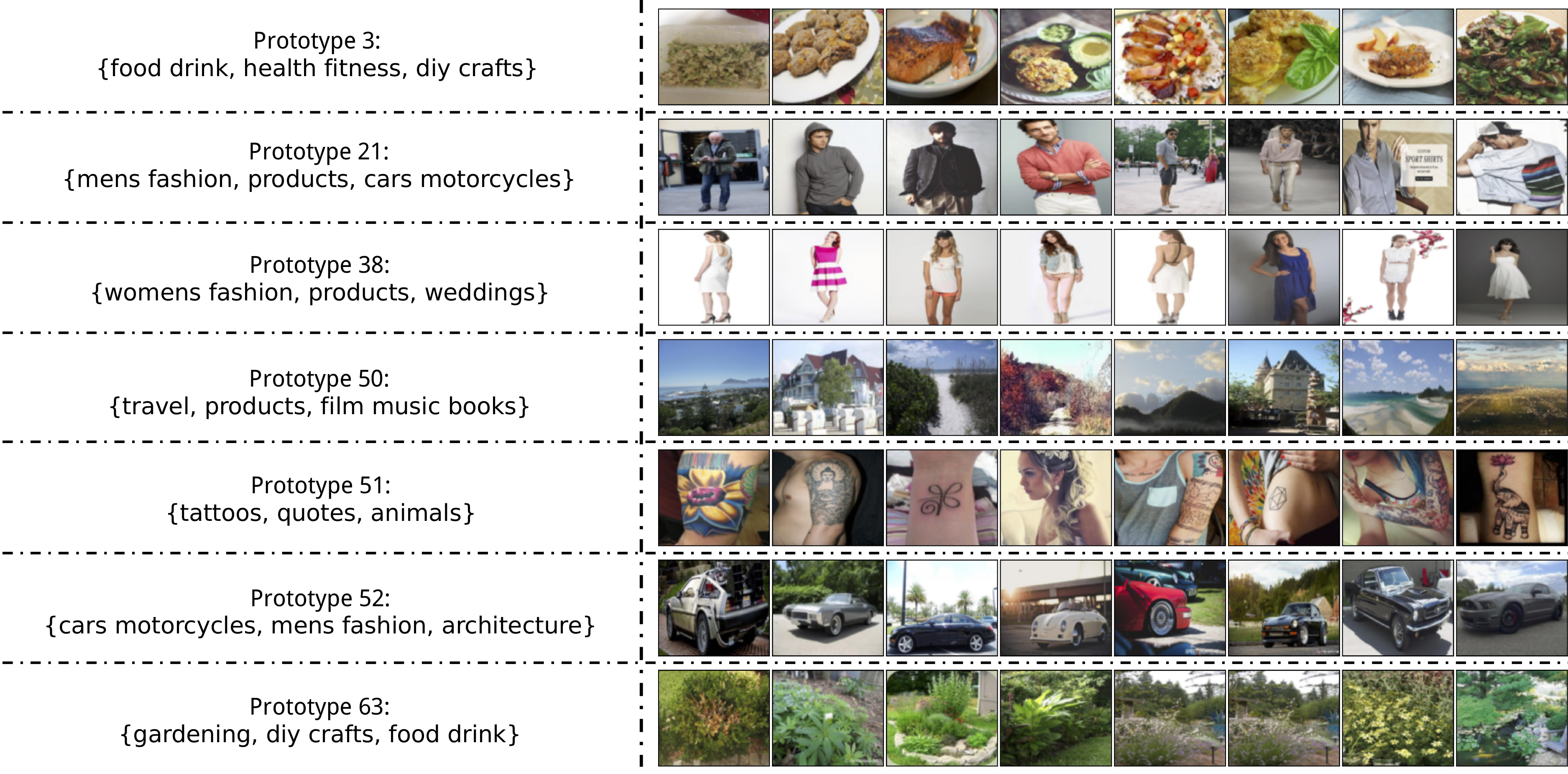}
  \centering
  \caption{Examples of prototype users.  The left panel shows the top three Pinterest categories that the prototype user most frequently posts images to, while the right panel shows representative images posted by individual users who are mapped to that prototype.}
  \label{fig:pintree_categories}
\end{figure*}

\section{Approach}\label{sec:approach}

Our goal is to predict the path of diffusion of an image in an online social network.  We first describe how we map individual users to a set of prototype users so that our proposed model can generalize to unseen users.  We then describe our deep recurrent Diffusion-LSTM network architecture for modeling and predicting image diffusion.

\subsection{Mapping individual users to prototypes} 

A diffusion path of an image encodes the sequence of users who shared it.  The set of all possible diffusion paths depends on both the users in the social network as well as the network structure (i.e., connectivity between users).  Since we want our model to be able to generalize to unseen users and unseen connections (i.e., be able to train and test with a different set of users in the social network), we map individual users to a fixed set of canonical prototype users who capture the general, shared characteristics of the individual users that are mapped to them (like in which categories they post, how popular they are). In this way, we can maintain the same prototypical users during training and testing, while ignoring subtle differences between the true users. Furthermore, when there is insufficient training data for each user, combining similar individual users into a single prototype user can enlarge the training data to produce a better prediction model.
%\color{red} Usually we are more interested in the characteristics of users (like in which categories they post, how popular they are) which user prototypes capture rather than the exact user itself. 
%\color{black}

%training data (number of connections) for the individual user is limited which might not be sufficient to learn the diffusion path prediction model. Whereas prototypical users will have much larger connections between them which will help us to train a better prediction model. }

%The possibilities in diffusion path prediction are each user's social friends in the network, which can be different during training and testing.
%since we want to be able to generalize to unseen users. To handle this discrepancy would be to group the users into a fixed set of prototypical users (e.g., via clustering) that is consistent during training and testing. We can then map each individual user to a particular prototype.

Specifically, we use $k$-means to cluster the social features (described in more detail below) of the users into $k=100$ prototype groups.\footnote{We set the number of prototypes to 100 to cover a broad range of users.  Empirically, we find that our algorithm is robust to a wide range of $k$ values.} The social feature of a prototype group is the average social feature of all of its cluster members.  An unseen user is mapped to the prototype (i.e., cluster center) whose social feature is most similar based on their euclidean distance. Figure~\ref{fig:pintree_categories} shows examples of prototype users.  The left panel shows the top three Pinterest categories that the prototype user most frequently posts images to, while the right panel shows representative images posted by individual users who are mapped to that prototype.  For example, prototype user 38 likes to share images of women's fashion, while prototype user 52 likes to share images of cars.

%To further process user social features, we first use K-means to find $1000$ user groups and then represent each user using Euclidean distances with respect to $1000$ user group centers. At last, for each of $1000$ social features, we subtract its mean and divide by its standard deviation. To obtain user group or prototype numbers, we use K-means again to cluster all users into $100$ unique groups. The number of prototypes is reduced to $100$ due to the difficult of the prediction task.

%It's important to include social category features because we use K-means to group all users into a fixed set of prototypical users. With social category features, prototypes returned from K-means will be more meaningful.

\begin{figure*}
\includegraphics[width=0.95\linewidth]{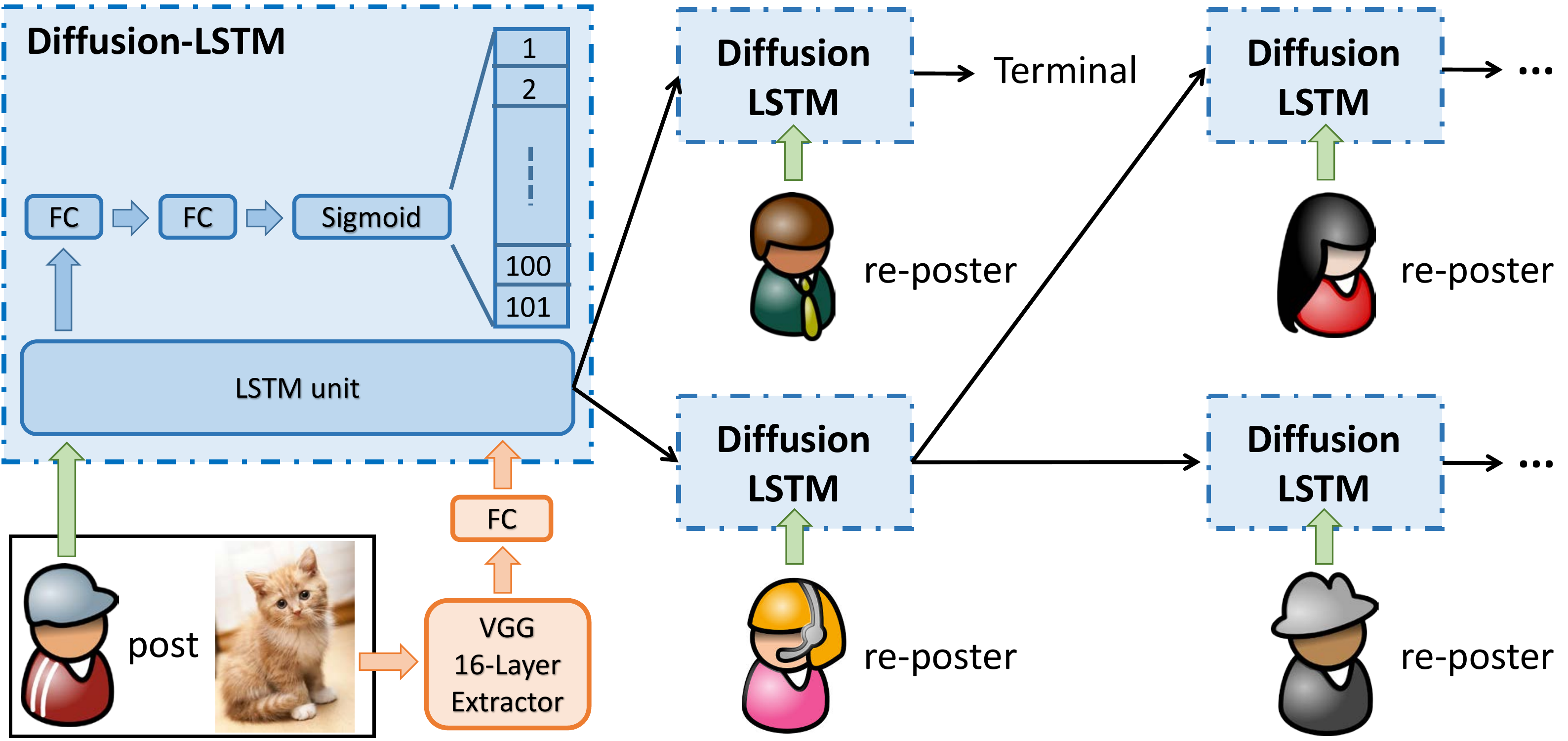}
  \centering
  \caption{The architecture of the Diffusion-LSTM model. The posted image goes through the VGG 16 network and the resulting image features are fed into the Diffusion-LSTM network as the initial memory $c_0$ through an FC layer. In each step, the LSTM-unit takes as input the current user's social features. The hidden vector output of the LSTM-unit $h$ goes through two FC layers and then a sigmoid function to get $101$ class probabilities (100 prototype users and 1 terminal class).  The Diffusion-LSTM cell in the dashed rectangle is repeated at successive tree nodes, with the only difference between the root node and children nodes being that the children nodes get the memory content $c$ and output $h$ from their parent node.
  \label{fig:FC-Tree-LSTM-FC}}
\end{figure*}
%The initial output $h_0$ is fixed to zero.

\subsection{Network architecture} 

The proposed architecture for modeling image diffusion through a social network is shown in Figure~\ref{fig:FC-Tree-LSTM-FC}.  The input to our model is an image $I$ posted by a user and the user's social information $S$, and the output is the entire diffusion tree $\{U_1, U_2, ..., U_n\}$ (sequence of re-posters) of the image.  Specifically, the initial diffusion path is conditioned on the input image and the social features of the root user. Then, in each ensuing timestep, our model takes the history of the previous posters' social features and the current poster's social features, and outputs the next re-poster (among the current poster's friends in the social network).  The model stores and updates a memory that records the characteristics of the users in the diffusion path so far. The process ends when the next predicted user is ``terminal''.

We next describe the image and social features, our Diffusion-LSTM model, and the loss function used to optimize the model.

%our model will take as input the image features (i.e., objects, attributes, style, etc.) as well as the current poster's social features (i.e., number of followers, number of likes, number of followings, etc.) and output the next re-poster (among the current poster's friends in the social network). The model will store and update a memory that records the characteristics of the users in the diffusion path so far. The process ends when the next predicted user is ``null''.

\subsection{Image and social features} 

The propagation path of an image in a social network should be a function of the image content as well as the users' social features.  For example, an interesting image is more likely to become viral if it is initially spread by a user who has many friends and followers.% \yj{GIVE SIMPLE EXAMPLE.} %Thus, we use both image and social features for modeling the diffusion path.

For image features, we take the VGG 16 network~\cite{simonyan-iclr2015} pretrained on ImageNet classification, and compute the FC7 activation feature (4096-D) for each image.  This feature encodes high-level semantics in the image (e.g., its objects/parts/attributes).

For social features, for each user, we compute the total number of followers of that user, the total number of users that the user follows, the total number of pins (shares) of all images shared by that user, and the total number of likes of all images shared by that user.  We also compute the probability distribution of the 38 Pinterest categories of all images posted by the user.  This captures the types of images that the user likes to share.  The 4-D aggregate counts and the 38-D category distribution are concatenated to create a single 42-D social feature.\footnote{While additional social factors such as gender, geographic location, age could also be used to enrich the feature, these factors are unfortunately missing for many of the users in the Pinterest dataset collected in~\cite{han-cosn2015}.}

\subsection{Diffusion-LSTM}

At any step along the diffusion path, whether an image will be further propagated will depend not only on the social influence of the current poster, but also on that of the previous posters (i.e., the history of the diffusion path taken thus far). This is an intrinsically recursive problem, and is well-suited to be modeled using Recurrent Neural Networks, which are ideal for analyzing sequences.  

Since we want to model long-range dependencies across many diffusion steps, we use an LTSM network~\cite{hochreiter-nc1997, gers-nc2000, gers-jmlr2002}, which is more robust to the vanishing gradient problem than vanilla RNNs~\cite{hochreiter-ijufks1998, benjio-tnn1994}. LTSMs have been successfully used for the related problem of image captioning (e.g.,~\cite{vinyals-cvpr2015}), where image features are combined with text features to recursively predict the next word in the image caption. In our case, the image features are combined with the current user's social features to recursively predict the next users who will share the image (recall Figure~\ref{fig:LSTM_overview} right).

%However, unlike in image captioning where a sequence of words form a chain, in our case, the sequence of users who share an image will form a tree (recall Fig.~\ref{fig:LSTM_overview} right).

%However, unlike in image captioning where a sequence of words form a chain, in our case, the sequence of users who share an image will form a tree (recall Fig.~\ref{fig:LSTM_overview} right).

\begin{figure*}
\includegraphics[width=1\linewidth]{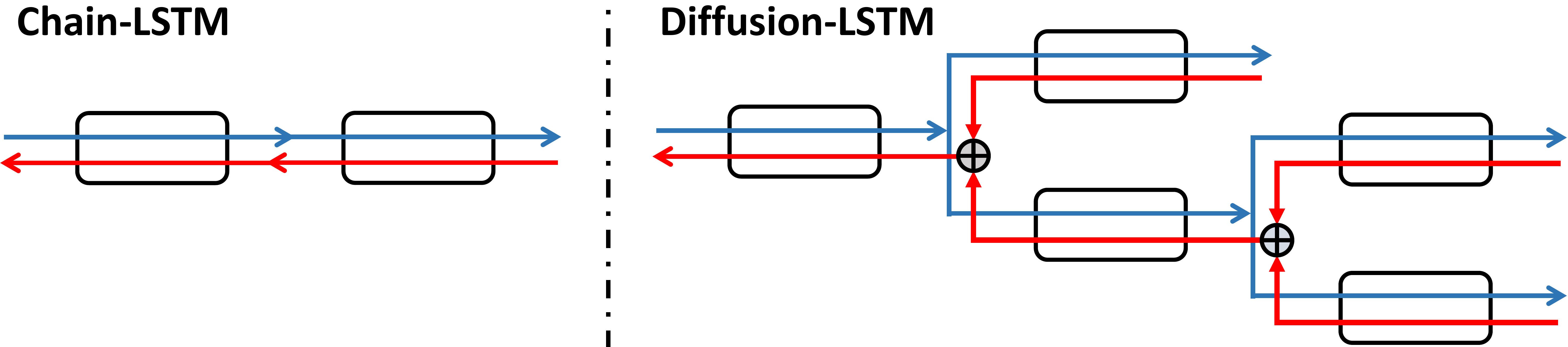}
  \centering
  \caption{The blue and red lines represent forward and backpropagation, respectively. (\textbf{Left}): A chain-structured LSTM network. Nodes from left to right are arranged according to time step $t$. (\textbf{Right}): The Diffusion-LSTM network. Nodes are arranged according to its parent-children relationship.  The \emph{sum} of the gradients of children is passed to the parent node during backpropagation.}
  \label{fig:LSTM_D-LSTM}
\end{figure*}

%children nodes directly. There is no clear time step definition here and the backpropagation of the Tree-LSTM network is implemented once per tree.
%To this end, we use the Tree-LSTM network proposed in~\cite{tai-arxiv2015, bowman-iccc2015, zhu-icml2015, zhang-arxiv2015}.
Our Diffusion-LSTM network for modeling image diffusion works exactly like a regular chain-structured LSTM network, with the only difference being the flow path of the memory cell $c$ and the hidden state $h$ following a tree. In a regular LSTM, the previous memory cell and hidden state from time step $t-1$ are passed to a single next node in time step $t$.  However, in our Diffusion-LSTM network, the previous memory cell and hidden state of a parent node from time step $t-1$ can be passed to multiple child nodes in time step $t$, as shown in Figure~\ref{fig:LSTM_D-LSTM}.  This means that during backpropagation, the gradient for a parent node will sum the gradients passed by all of its child nodes. %\yj{NEED TO SAY MORE ABOUT HOW THIS CHANGES THE COMPUTATION VS A REGULAR LSTM.}

Formally, using the subscripts $ch$ and $p$ for ``child'' and ``parent'' respectively, the Diffusion-LSTM parent-child transition equations are expressed as follows:
\begin{align}
f_{ch} &=\sigma(W^f_{x}x_{ch}+W^f_{h}h_{p}+b^{f}) \nonumber \\
i_{ch} &=\sigma(W^i_{x}x_{ch}+W^i_{h}h_{p}+b^{i}) \nonumber \\
o_{ch} &=\sigma(W^o_{x}x_{ch}+W^o_{h}h_{p}+b^{o}) \nonumber \\
c_{ch} &=f_{ch}\circ c_{p}+i_{ch}\circ \tanh(W^c_{x}x_{ch}+W^c_{h}h_{p}+b^{c}) \nonumber \\
h_{ch} &=o_{ch}\circ \tanh(c_{ch})
\label{eq:LSTM_equations}
\end{align}
where $x_{ch}$ is the input social feature of the user at the current step, $W_x$, $W_h$ and $b$ are the weight matrices and bias vectors to be optimized, and $\circ$ denotes elementwise multiplication. $f_{ch}$ denotes the forget gate vector (for remembering old information), $i_{ch}$ denotes the input gate vector (for acquiring new information), $o_{ch}$ denotes an output gate vector (as an output candidate), $c$ denotes the memory cell, and $h$ denotes the hidden state.

Our Diffusion-LSTM model consists of the regular LSTM unit followed by two FC layers and one sigmoid layer afterwards to predict one of the $k=100$ prototypical users or the ``terminal'' class. To reduce overfitting and increase the nonlinearity of the model, we add one dropout layer after the LSTM unit, and one ReLU layer and one dropout layer between two FC layers.

%\yj{WE NEED TO SAY WHY THIS MAKES SENSE!!!}
We project the image features into the initial root parent memory $c_p^{root}$ through an FC layer.  This conditions the initial history of the propagation path with the contents of the image. Alternatively, one could concatenate a user's social features with the image features, but this produces inferior results.  The initial hidden state $h_0$ is simply set to all zero.

%Instead of directly concatenating image features to social features for the root node, we feed image features to $c_p^{root}$ mainly because we want to make the inputs to LSTM at each node to be ``homogeneous'', i.e., all receiving social features. Also, feed image features as input instead of memory might have the danger of image features not propagating further down the tree. The initial hidden state $h_0$ is simply set to all zero. %At every tree node, while making predictions, besides $100$ prototype classes, we add an additional terminal class as the terminating signal.
% between image features and the Diffusion-LSTM unit

\subsection{Loss function for modeling image diffusion}

The diffusion path of an image can be broken-down into a sequence of classification problems, where the predicted classes for a given timestep indicate the users who will reshare the image in the ensuing timestep (i.e., next depth level of the social network tree).  Furthermore, since the number of users who reshare the image can range anywhere from no user (zero) to $k$ different users, we model the diffusion prediction for a given timestep and tree node as a collection of $k+1$ binary classification problems (where the $k+1$'th class denotes the terminal class). Thus, we use the weighted multi-class binary cross entropy loss to optimize our model:
\begin{align}
L = \sum_{u=1}^{k+1} -\frac{1}{m} \sum_{i=1}^m (w^p_{u} t_{u,i} \log(o_{u,i}) + w^n_{u} (1 - t_{u,i}) \log(1 - o_{u,i}))
\label{eq:loss}
\end{align}
where $u$ indexes the classes (prototype users and terminal class), $i$ indexes the training instances (i.e., all nodes in all training trees), $o_{u,i}$ is the predicted probability of instance $i$ being in class $u$, and $t_{u,i}$ is the ground-truth binary indicator indicating whether instance $i$ is in class $u$.  

Since the number of positive vs. negative instances in a class can be highly imbalanced, we use the weights $w^p_u$ and $w^n_u$ to balance their influence: $w^p_u = N_{un}/(N_{un} + N_{up})$, $w^n_u = N_{up}/(N_{un} + N_{up})$ and $N_{un}$ ($N_{up}$) is the number of negative (positive) ground-truth instances for class $u$.  Without these balancing weights, the loss function can be dominated by the dominant (positive or negative) instances.

%To train the model, we will use the ground-truth propagation paths in the training data.
%For each class, the prediction task is treated as an individual binary classification problem and the loss for each node is measured by the weighted multi-class binary cross entropy between the targets and the outputs, defined in the technique approach section. The overall loss is then described as the mean of all node losses.

%For a multi-class binary classification problem, with unbalanced numbers of positive ($1$) targets and negative ($0$) targets of each class, we define the weighted multi-class binary cross entropy:
%\begin{align}
%loss(\mathbf{o}, \mathbf{t}) = -\frac{1}{n} \sum_i (w^p_i t_i \log(o_i) + w^n_i (1 - t_i) \log(1 - o_i))
%\label{eq:loss}
%\end{align}
%where $w^p_i = N_{in}/(N_{in} + N_{ip})$, $w^n_i = N_{ip}/(N_{in} + N_{ip})$ and $N_{in}$ ($N_{ip}$) is the number of negative (positive) targets in the ground truth for class $i$. Outputs $\mathbf{o}$ describe multi-class probabilities which are between $0$ and $1$, and targets $\mathbf{t}$ represent multi-class ground-truths that are either $0$ or $1$.

\section{Experiments}

In this section, we evaluate the accuracy of our model's image diffusion path prediction, and compare to several baselines that either use only image or social features, or lack memory.  We also show qualitative examples of our model's generated diffusion trees.%, and measure their similarity to the ground-truth trees.

 %detailed results of comparing our Diffusion-LSTM model with several designed baselines.  Also, we use the trained Tree-LSTM model to generate diffusion trees directly and measure their affinity with original pintrees quantitatively.

%, number of followers in twitter, number of followings in twitter and number of tweets
%image files,
\subsection{Pinterest dataset} To analyze image content diffusion, we use the anonymized Pinterest data of~\cite{han-cosn2015}, which was collected for 44 days in 2013. The dataset consists of image propagation data (pintree id, sender id, receiver id), user data (user id, number of pins (shares), number of followers, number of followings, number of likes, gender, locale, country), and image data (pintree id, Pinterest category, source, number of likes, and time stamp). For each image, its entire propagation path is available. Overall, the dataset contains more than 340K diffusion paths (Pinterest trees or ``pintrees'') shared by 1M users, which is randomly divided into training, validation, and test sets with size ratio $6:1:1$.  The mean pintree size is $4.48$ with standard deviation $11.67$, the mean pintree width is $3.15$ with standard deviation $6.85$, and the mean pintree depth is $1.66$ with standard deviation $1.05$.
\begin{table*}
   \caption{Average Precision for per-node diffusion prediction. Our full model (last row) produces the best results compared with alternative baselines that rely only on either image or social features, or do not encode the image's posting history. See text for details.}
  \centering
  \begin{tabular}{lcccccc}
    \hline
    Model & Social & Image & Memory & AP$_{101}$ & mAP$_{1-100}$ & mAP  \\
    \hline
    Random Weights & Yes & Yes & Yes & 0.803 & 0.005 & 0.013 \\
    %FC & Yes  & Yes & No & 0.902 & 0.145 & 0.152 \\
    FC & Yes  & Yes & No & 0.899 & 0.126 & 0.133 \\
    %FC & Yes  & No & No & 0.902 & 0.145 & 0.152 \\
    %FC & No  & Yes & No & 0.835 & 0.039 & 0.047 \\
    Diffusion-LSTM & No & Yes & Yes & 0.883 & 0.087 & 0.095 \\
    Diffusion-LSTM & Yes & No & Yes & \textbf{0.932} & 0.206 & 0.213 \\
    Diffusion-LSTM & Yes & Yes & Yes & \textbf{0.932} & \textbf{0.214} & \textbf{0.221} \\	
    \hline
    \end{tabular}%  Image features improve prediction performance for users who post to multiple categories ($N>1$).
  \label{mAP_table}
\end{table*}

For social features, we use number of pins, number of followers, number of followings, and number of likes, because all other user data have more than $78\%$ data missing, while those four social features have less than $1\%$ data missing (which we simply fill with their mean values). As explained in Sec.~\ref{sec:approach}, we expand the user social features with the $38$ Pinterest category distributions, which can be summarized from all images shared by a user.

%We consider the following image content factors: (i) image metadata information such as its Pinterest category (e.g., "food", "animal"), and (ii) image featuresfrom low-level (e.g., color, texture, brightness) to semantic-level features (e.g., image styles, objects identified in an image) computed using the pretrained VGG 16-layer network.
%For social factors, we study four social features of the posters/receivers, including the number of pins, the number of followers, the number of followings and the number of likes. Also, we expand user social features with additional $38$ category features, summarized from all posts by users, which describe corresponding category probability distributions.
%It's important to include social category features because we use K-means to group all users into a fixed set of prototypical users. With social category features, prototypes returned from K-means will be more meaningful. In Fig.~\ref{fig:pintree_categories}, we illustrate some chosen images, closest to the prototype center in the feature space, for certain prototypes. The left panel of Fig.~\ref{fig:pintree_categories} indicates top three popular Pinterest categories for each prototype.

%The Diffusion-LSTM with random weights is also compared with our approach.  

%The FC model is trained with a fixed learning rate $0.01$, while
\subsection{Implementation details} We train our Diffusion-LSTM model with an initial learning rate of $0.2$, and then lower the learning rate to $0.02$ when validation loss stops decreasing. We train both our model and all baselines until full convergence (30-60 epochs depending on the model). We optimize the weights of the model using backpropagation through time.  We normalize each social feature to the range $[0, 1]$ by first applying the log function and then dividing by the maximum value. We divide each Pinterest category distribution feature by $4$ to make it have roughly the same standard deviation as each non-category social features.  For the VGG-16 image features, we subtract each dimension by its mean and divide by its standard deviation.

\subsection{Baselines} We compare our Diffusion-LSTM model with several baselines:

%To evaluate the Tree-LSTM model, we design and test several baselines, including the FC model trained with either social features or image features and the Tree-LSTM model trained with either social or image features. FC models are chosen to study the importance of memory, while the Tree-LSTM model with either social or image features is used to explore the role of each in the prediction task.

\emph{Diffusion-LSTM with random weights}: The same architecture as our Diffusion-LSTM model but with random weights. The initial memory $c_0$ is projected from image features at the beginning. This is a sanity check baseline to measure chance performance.%However, we don't do any backpropagation to update weights and initial weights are totally randomized.

\emph{FC model with image and social features}: This baseline consists of three fully connected (FC) layers.  One ReLU layer and one dropout layer follow each of the first two FC layers.  The last FC layer is followed by a sigmoid layer to convert the output values into probabilities.  This baseline is chosen to study the importance of memory for content diffusion path prediction.

%This model is trained on every tree node but updates its weights every single pintree. Only social features are used for this baseline.

%\textbf{FC model with only social features}: As a baseline model, this model mainly consists of three fully connected (FC) layers, with each of two front FC layers followed by one ReLU layer and one dropout layer and the last FC layer followed by the sigmoid layer to directly convert output values into probabilities. This model is trained on every tree node but updates its weights every single pintree. Only social features are used for this baseline.

%\textbf{FC model with only image features}: The same architecture with the first baseline but with only image features used as the input.

\emph{Diffusion-LSTM with only social features}: The same architecture as our Diffusion-LSTM model but with only social features. No image features are used for this baseline, so initial memory $c_0$ is simply set to all zero.  This baseline measures the role of social features for content diffusion path prediction.
%used as the input of Tree-LSTM unit

\emph{Diffusion-LSTM with only image features}: The same architecture as our Diffusion-LSTM model but with only image features. The initial memory $c_0$ is also set to all zero for this baseline.  Since social features are not used, we instead use the (same) image features as input in each time step.  This baseline measures the role of image features for content diffusion path prediction.  

Note that we cannot compare with Tree-LSTMs~\cite{tai-acl2015,zhu-icml2015,zhang-naacl2016} because it is not possible to apply them to our problem setting (i.e., top-down tree prediction/generation) due to their bottom-up nature.

\subsection{Per-node diffusion prediction accuracy}

We first evaluate image diffusion prediction by comparing our predicted users with the ground-truth users for each node in a tree.  We evaluate per-category average precision (AP), by varying the threshold in the prediction confidences.  %\yj{Note that for this evaluation,}
Following the standard practice in evaluating RNNs~\cite{milkov-interspeech2010}, we do not generate the tree (we will do this in the next section), but instead use the ground-truth path up to timestep $t-1$ for a tree node and evaluate whether the predicted users in timestep $t$ match the ground-truth (and do this for all $t$ and all nodes).

%For each prototype class, by outputting ground-truth values and corresponding confidence values, we calculate its average precision (AP), defined as the area under the precision-recall curve. Then, we obtain the mean average precision (mAP) by averaging the AP of associated classes.

%Testing results of two models with different training schemes have been shown in Table~\ref{mAP_table}, where AP$_{101}$ stands for the terminal class AP, mAP$_{1-100}$ stands for the prototype class mAP and mAP represents the mAP for all $101$ classes.

Table~\ref{mAP_table} shows the per-node prediction results. AP$_{101}$ is the AP for the terminal class, while mAP$_{1-100}$ is the mean AP for all prototype user classes. mAP is the mean AP computed over all $101$ classes.  First, we can see that all baselines outperform the random weights baseline, which produces a very low mAP showing the difficulty of the task. The reason why the AP$_{101}$ for random weights is so high is because there are many leaf nodes (terminal class) in the database. Specifically, the terminal class has a $1:4$ ratio of non-leaf vs. leaf (total of $275,595$ non-leaf and $1,128,372$ leaf). On the other hand, for the $1$-$100$ prototype user classes, the negative vs. positive instance ratio is around $200:1$ (an average total of $1,397,230$ negatives and $6,736$ positives). The large difference in target ratios make the prediction task for the prototype classes much more difficult than the terminal class.
%$275595$ negative ($0$) targets and $1128372$ positive ($1$) targets, with the

Second, we see that social features alone produce better prediction results than image features alone.  This corroborates many previous findings (e.g.,~\cite{deza-cvpr2015,khosla-www2014,totti-cws2014}), which showed that social features are more important than image features for content popularity prediction.  In our case, we are predicting the diffusion path, which is a much more difficult problem, and when no social features are used, it is impossible to say precisely how an image will be propagated through a social network. Third, our Diffusion-LSTM model performs much better than the FC model, indicating the crucial role of storing the image posting history in memory for predicting the next re-posting users.

\begin{figure}[t]
\includegraphics[width=0.8\linewidth]{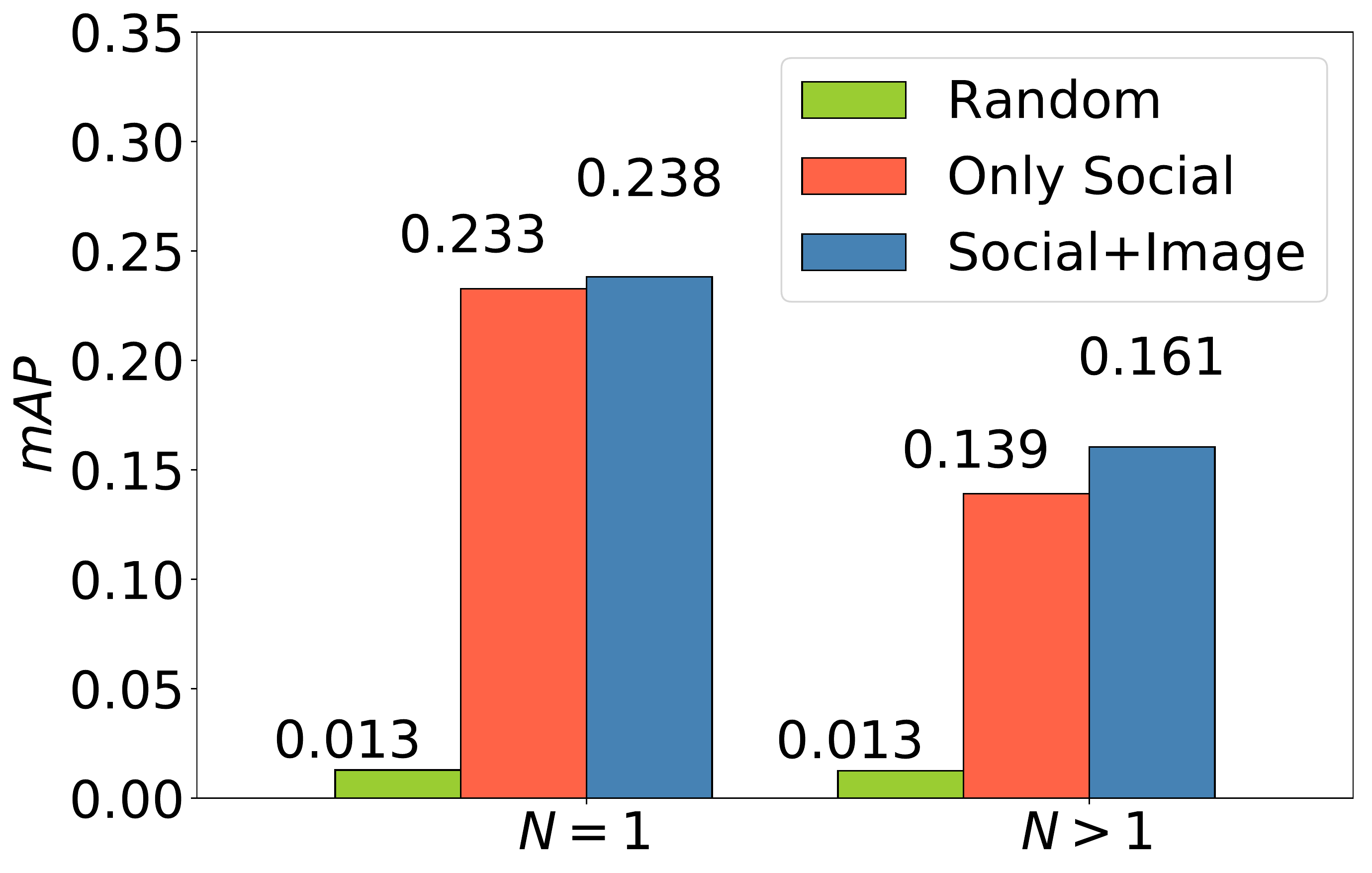}
  \centering
  \caption{Single-category (left) vs.~multi-category posters (right). N represents the number of categories a user post to. In each case, our full model (blue) is compared with the baseline which uses only social features (red) and the baseline which uses random weights (green). Our full model demonstrates significant improvement in prediction performance over these baselines for users who post to multiple categories.}
  \label{fig:LSTM_compare}
  \vspace*{-0.1in}
\end{figure}

\begin{table*}
   \caption{Quantitative evaluation comparing our generated pintrees to ground-truth pintrees.  Our full model (last row) generates pintrees that are closer to ground-truth than alternative baselines.}
  \centering
  \begin{tabular}{lccccc}
    \hline
    Model & Social & Image & Memory & Depth MAE & HI \\
    \hline
    Chance Performance & Yes & Yes & Yes & 2.44 & 0.007 \\
    FC & Yes  & Yes & No & 2.46 & 0.278\\
    Diffusion-LSTM & Yes & No & Yes & 1.98 & 0.701\\
    Diffusion-LSTM & Yes & Yes & Yes & \textbf{1.85} & \textbf{0.708}\\	
    \hline
  \end{tabular}
  \label{test_th_evaluation}
\end{table*}

Finally, combining social features and image features produces the best results, improving overall mAP by $0.8\%$ over the baseline Diffusion-LSTM model using only social features.  The main reason why the improvement is only $0.8\%$ is because most ($88.2\%$) users only post images in one Pinterest category instead of multiple categories.  This means that those users are likely to always share similar images (e.g., of cats) to similar prototypical users (e.g., cat lovers).  (Recall that we also use the probability distribution of the 38 Pinterest categories of all images posted by the user as part of the social features.)  On the other hand, for users who post images of multiple Pinterest categories (e.g., cats, cars, travel, food), the image content will influence who among his/her followers will repost the image.  To study this in more detail, we further break down prediction accuracy based on whether a user posts only to a single category versus multiple categories.  Indeed, Figure~\ref{fig:LSTM_compare} shows that image features yield only a slight improvement over social features for users who post only to a single category ($N=1$), but significantly improve prediction performance for users who post to multiple categories ($N>1$).

\subsection{Tree generation accuracy}

We next use our trained Diffusion-LSTM model to generate pintrees directly.  We compare the generated tree with the ground-truth tree, given the same starting user and posted image. We compare the depth of the trees (which evaluates how well our model predicts the terminal class) and the user's preferred Pinterest category distributions averaged across all non-root nodes (users) in the tree (which measures how well our model predicts the prototype users).  We do not consider the root node since it is already given during generation.  We compute the mean absolute error (MAE) for depth, and histogram intersection for the Pinterest category distribution: $HI(\mathbf{d}_1, \mathbf{d}_2) = \sum_i min(d_{1i}, d_{2i})$, where $\mathbf{d}_1$, $\mathbf{d}_2$ are two normalized Pinterest category distributions ($38$-D histograms).  

To generate a tree, given the features for a user and associated image, we recursively make $101$ class (100 prototype users or terminal class) predictions.  If a node is predicted as ``terminal'', we stop; otherwise, we continue generating the tree.  When generating a tree, we set the terminal class prediction threshold to be $0.5$, while the prototype class prediction thresholds are chosen based on validation data using depth MAE.  Due to the unequal number of ground-truth pintrees per depth, we evaluate the MAE for each tree depth separately and then average across depths. We truncate any pintree if its depth $>10$ or size $>150$ or width $>100$ to avoid degenerate trees that never stop growing (otherwise, they induce infinite error in our evaluation).  99.9\% of the pintrees in the training set fall under these criteria.
%The thresholds for the baselines are chosen in the same way.

%We compare the size, width, and depth of the trees, as well as the user's preferred Pinterest category distributions (averaged across all nodes (users) in the tree).  We compute mean absolute error (MAE) when comparing size, width, and depth, and the histogram intersection (HI) when comparing the Pinterest category distribution: $HI(\mathbf{d}_1, \mathbf{d}_2) = \sum_i min(d_{1i}, d_{2i})$, where $\mathbf{d}_1$, $\mathbf{d}_2$ are two Pinterest category distributions ($38$-D histogram vectors).  To generate a tree, given the features for a user and associated image, we recursively make $101$ class (100 prototype users or terminal class) predictions.  If a node is predicted as ``terminal'', we stop; otherwise, we continue generating the tree.  \yj{When generating a tree, we set the terminal class prediction threshold to be $0.5$, while the prototype class prediction thresholds are chosen based on validation data using depth MAE.}  The thresholds for the baselines are chosen in the same way.

%\begin{align}
%HI(\mathbf{d}_1, \mathbf{d}_2) = \sum_i min(d_{1i}, d_{2i})
%\label{eq:histogram_intersection}
%\end{align}

\begin{figure*}
\includegraphics[width=1\linewidth]{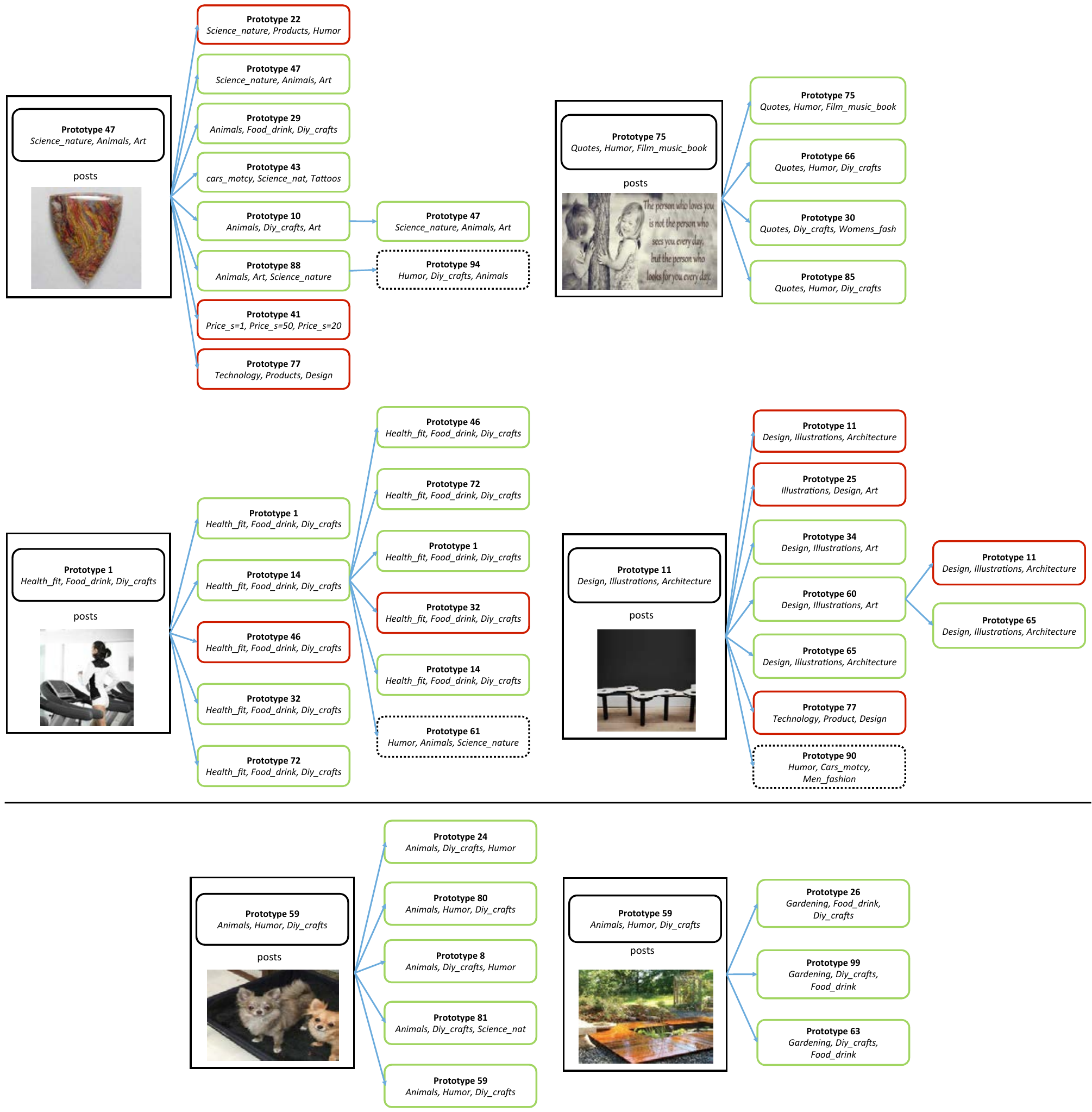}
  \centering
  \caption{
  \textbf{(Top)} For a given image and its user, we generate the entire pintree showing the image propagation path through the different users.  A user is shown in: 1) green box if the predicted path to that user matches the ground-truth path; 2) red box if it does not match the ground-truth; 3) dotted box if the path is present in the ground-truth pintree but not in our generated pintree. We also show the top 3 Pinterest categories for each user. Our predicted pintrees closely resemble the ground-truth pintrees. Note that while a self-loop can be observed at the prototype level (e.g., as seen in the bottom-left `Prototype 1' social tree generation example), it still represents a mapping between different individual users since different individual users can be mapped to the same user prototype.  \textbf{(Bottom)} Comparison of generated pintrees when the same prototype user posts different images. The generated tree depends on both the user's social features and image content. 
  %\yj{Two generated pintrees with given images (from the testing dataset) and corresponding specific users.
%Each pintree is visualized to start with the given image and the prototype number which the initial user belonging to.
%The image in the left upper conner of each pintree is the initial post and numbers right after circles stand for prototype numbers which users belonging to. Right after prototype numbers, we randomly pick up and show four images also posted by that prototype members. Below generated images, corresponding diffusion groundtruths are also presented.}
}
  \label{fig:pintree_visu}
\end{figure*}

Table~\ref{test_th_evaluation} shows the results.  Our generated trees are more similar to the ground-truth trees than the FC baseline, which again shows the importance of memory, and the Diffusion-LSTM baseline that lacks image features (the one that lacks social features does much worse, similar to per-node prediction).  All methods outperform chance performance (sampling trees according to the training data distribution), which shows that learning is important for diffusion prediction. Figure~\ref{fig:pintree_visu} (top) shows visualizations of pintrees generated by our Diffusion-LSTM model.  Our predicted pintrees are quite similar to the ground-truth pintrees.  Even for the mispredicted users (red boxes), their Pinterest categories are very close to the ground-truth users' categories.  Interestingly, for our top-left generated pintree, our model predicts that a user from prototype 47 will share the image from a user from prototype 10 who initially shared the image from another user from prototype 47. This shows the importance of memory as our model can remember that the image was initially shared by a user from prototype 47 and uses this to make the correct prediction. 

Finally, in Figure~\ref{fig:pintree_visu} (bottom), we show how the generated pintrees change when the same prototype user posts different types of images. When a user who posts frequently to the animal category posts a gardening image (Figure~\ref{fig:pintree_visu} (bottom) right), it is shared by users who post frequently to the gardening category (indicating image plays a role).  Social features still play a role, as can be seen by the common Pinterest categories between the original user and re-sharers (Diy\_crafts is in common in both Figure~\ref{fig:pintree_visu} (bottom) left and right). This indicates image propagation is dependent on both image content and user social features.

\section{Conclusion}% and Future Work

We presented a deep recurrent network for generating the diffusion path of an image through a social network.  By keeping track of the posting history of an image, our Diffusion-LSTM outperforms alternative baselines that lack memory or use only image or social features. We further demonstrated our model's capability of generating meaningful pintrees.

Although we mainly focused on Pinterest data, the proposed model is general and can be applied to different social network datasets. To further boost prediction performance, we could fine-tune the VGG 16 network, since the ImageNet dataset and Pinterest dataset may possess different characteristics (Pinterest images are more likely to be ``artsy'').  Finally, with sufficient training data for each user, it may also be possible to train the Diffusion-LSTM to directly predict the specific friends who will share the image, rather than predicting prototype users.

\section*{Acknowledgements} This work was supported in part by GPUs donated by NVIDIA and the research fund of Hanyang University (HY-2017-N).

\bibliographystyle{ACM-Reference-Format}
\bibliography{bibs}

\end{document}